\def\year{2022}\relax
\documentclass[letterpaper]{article} 
\usepackage{aaai22}  
\usepackage{times}  
\usepackage{helvet}  
\usepackage{courier}  
\usepackage[hyphens]{url}  
\usepackage{graphicx} 
\usepackage{natbib}  
\usepackage{caption} 
\DeclareCaptionStyle{ruled}{labelfont=normalfont,labelsep=colon,strut=off} 
\usepackage{algorithm}
\usepackage{algorithmic}
\usepackage{amsfonts}
\usepackage{amsmath}

\usepackage{newfloat}
\usepackage{listings}
\floatstyle{ruled}
\newfloat{listing}{tb}{lst}{}
\floatname{listing}{Listing}
\usepackage{multirow}

\title{FastFlow: Unsupervised Anomaly Detection and Localization \\ via 2D Normalizing Flows}
\author{
    Jiawei Yu\textsuperscript{\rm 1}\equalcontrib,
    Ye Zheng\textsuperscript{\rm 2,3}\equalcontrib,
    Xiang Wang\textsuperscript{\rm 1},
    Wei Li\textsuperscript{\rm 1},
    Yushuang Wu\textsuperscript{\rm 4},
    Rui Zhao\textsuperscript{\rm 1},
    Liwei Wu\textsuperscript{\rm 1}
}
\affiliations{
    \textsuperscript{\rm 1} SenseTime Research\\
    \textsuperscript{\rm 2}Institute of Computing Technology, Chinese Academy of Sciences\\
    \textsuperscript{\rm 3}University of Chinese Academy of Sciences
    \textsuperscript{\rm 4}The Chinese University of Hong Kong, Shenzhen\\
     
}

\usepackage{bibentry}

\begin{document}

\maketitle

    




\begin{abstract}
Unsupervised anomaly detection and localization is crucial to the practical application when collecting and labeling sufficient anomaly data is infeasible. Most existing representation-based approaches extract normal image features with a deep convolutional neural network and characterize the corresponding distribution through non-parametric distribution estimation methods. The anomaly score is calculated by measuring the distance between the feature of the test image and the estimated distribution. However, current methods can not effectively map image features to a tractable base distribution and ignore the relationship between local and global features which are important to identify anomalies. To this end, we propose FastFlow implemented with 2D normalizing flows and use it as the probability distribution estimator. Our FastFlow can be used as a plug-in module with arbitrary deep feature extractors such as ResNet and vision transformer for unsupervised anomaly detection and localization. In training phase, FastFlow learns to transform the input visual feature into a tractable distribution and obtains the likelihood to recognize anomalies in inference phase. Extensive experimental results on the MVTec AD dataset show that FastFlow surpasses previous state-of-the-art methods in terms of accuracy and inference efficiency with various backbone networks. Our approach achieves 99.4\% AUC in anomaly detection with high inference efficiency.


\end{abstract}

\section{Introduction}
\label{section:intro}


The purpose of anomaly detection and localization in computer vision field is to identify abnormal images and locate abnormal areas, which is widely used in industrial defect detection (Bergmann et al. 2019, 2020), medical image inspection (Philipp Seeböck et al. 2017), security check (Akcay, Atapour-Abarghouei, and Breckon 2018) and other fields. However, due to the low probability density of anomalies, the normal and abnormal data usually show a serious long-tail distribution, and even in some cases, no abnormal samples are available. The drawback of this reality makes it difficult to collect and annotate a large amount of abnormal data for supervised learning in practice. Unsupervised anomaly detection has been proposed to address this problem, which is also denoted as \textit{one-class classification} or \textit{out-of-distribution detection}. That is, we can only use normal samples during training process but need to identify and locate anomalies in testing.

\begin{figure}[t]
\centering
\includegraphics[width=0.97\columnwidth]{ 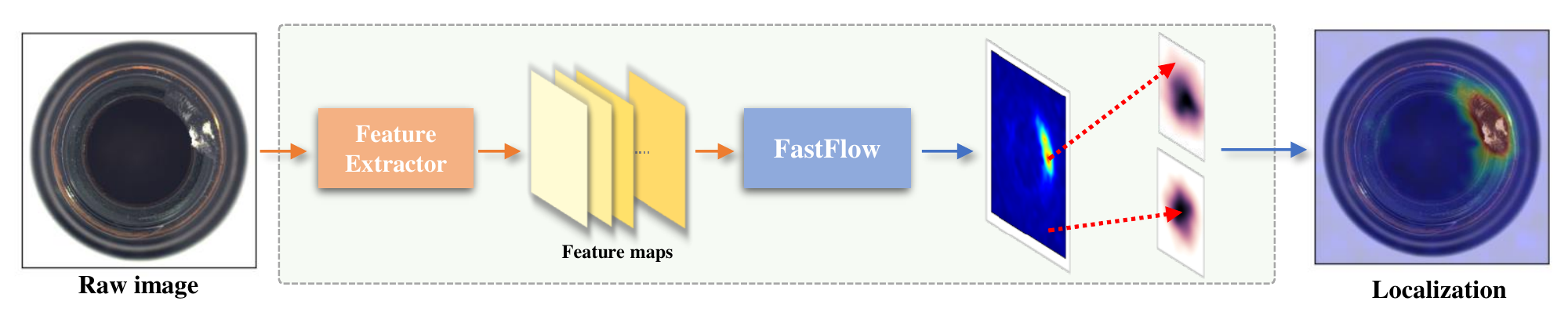}
\caption{An example of the proposed FastFlow. FastFlow transforms features of the input image from the original distribution to the standard normal distribution. The features of the normal area in the input image fall in the center of the distribution, while the abnormal features are far away from the center of the distribution.}
\label{fig:story}
\end{figure}

One promising method in unsupervised anomaly detection is using deep neural networks to obtain the features of normal images and model the distribution with some statistical methods, then detect the abnormal samples that have different distributions~\cite{bergman2020classification,rippel2021modeling,yi2020patch,cohen2020sub,defard2020padim}.
Following this methodology, there are two main components: the feature extraction module and the distribution estimation module. 

To the \textit{distribution estimation module}, previous approaches used the non-parametric method to model the distribution of features for normal images. For example, they estimated the multidimensional Gaussian distribution~\cite{li2021cutpaste,defard2020padim} by calculating the mean and variance for features, or used a clustering algorithm to estimate these normal features by normal clustering~\cite{reiss2021panda,roth2021towards}. Recently, some works~\cite{rudolph2021same,gudovskiy2021cflow} began to use normalizing flow~\cite{kingma2018glow} to estimate distribution. Through a trainable process that maximizes the log-likelihood of normal image features, they embed normal image features into standard normal distribution and use the probability to identify and locate anomalies. However, original one-dimensional normalizing flow model need to flatten the two-dimensional input feature into a one-dimensional vector to estimate the distribution, which destroys the inherent spatial positional relationship of the two-dimensional image and limits the ability of flow model. In addition, these methods need to extract the features for a large number of patches in images through the sliding window method, and detect anomalies for each patch, so as to obtain anomaly location results, which leads to high complexity in inference and limits the practical value of these methods. To address above problems, we propose the FastFlow which extend the original normalizing flow to two-dimensional space. We use fully convolutional network as the subnet in our flow model and it can maintain the relative position of the space to improve the performance of anomaly detection. At the same time, it supports the end-to-end inference of the whole image and directly outputs the anomaly detection and location results at once to improve the inference efficiency.

To the \textit{feature extraction module} in anomaly detection, besides using CNN backbone network such as ResNet~\cite{he2016deep} to obtain discriminant features, most of the existing work~\cite{defard2020padim,reiss2021panda,rudolph2021same,gudovskiy2021cflow} focuses on how to reasonably use multi-scale features to identify anomalies at different scales and semantic levels, and achieve pixel-level anomaly localization through sliding window method. The importance of the correlation between global information and local anomalies~\cite{yan2021learning,wang2021glancing} can not be fully utilized, and the sliding window method needs to test a large number of image patches with high  computational complexity. To address the problems, we use FastFlow to obtain learnable modeling of global and local feature distributions through an end-to-end testing phase, instead of designing complicated multi-scale strategy and using sliding window method. We conducted experiments on two types of backbone networks: vision transformers and CNN. Compared with CNN, vision transformers can provide a global receptive field and make better use of global and local information while maintaining semantic information in different depths. Therefore, we only use the feature of one certain layer in vision transform. Replacing CNN with vision transformer seems trivial, but we found that performing this simple replacement in other methods actually degrade the performance, but our 2D flow achieve competitive results when using CNN. Our FastFlow has stronger global and local modeling capabilities, so it can better play the effectiveness of the transformer. 

As shown in Figure~\ref{fig:story}, in our approach, we first extract visual features by the feature extractor and then input them into the FastFlow to estimate the probability density. In training stage, our FastFlow is trained with normal images to transform the original distribution to a standard normal distribution in a 2D manner. In inference, we use the probability value of each location on the two-dimensional feature as the anomaly score.

To summarize, the main contributions of this paper are:
\begin{itemize}
\item We propose a 2D normalizing flow denoted as FastFlow for anomaly detection and localization with fully convolutional networks and two-dimensional loss function to effectively model global and local distribution.
\item We design a lightweight network structure for FastFlow with the alternate stacking of large and small convolution kernels for all steps. It adopts an end-to-end inference phase and has high efficiency.
\item The proposed FastFlow model can be used as a plug-in model with various different feature extractors. The experimental results in MVTec anomaly detection dataset~\cite{bergmann2019mvtec} show that our method outperforms the previous state-of-the-art anomaly detection methods in both accuracy and reasoning efficiency.
\end{itemize}

\section{Related Work}
\subsection{Anomaly Detection Methods}
Existing anomaly detection methods can be summarized as reconstruction-based and representation-based methods. Reconstruction-based methods~\cite{bergmann2019mvtec,gong2019memorizing,perera2019ocgan} typically utilize generative models like auto-encoders or generative adversarial networks to encode and reconstruct the normal data. These methods hold the insights that the anomalies can not be reconstructed since they do not exist at the training samples. Representation-based methods extract discriminative features for normal images~\cite{ruff2018deep,bergman2020classification,rippel2021modeling,rudolph2021same} or normal image patches~\cite{yi2020patch,cohen2020sub,reiss2021panda,gudovskiy2021cflow} with deep convolutional neural network, and establish distribution of these normal features. Then these methods obtain the anomaly score by calculating the distance between the feature of a test image and the distribution of normal features. The distribution is typically established by modeling the Gaussian distribution with mean and variance of normal features~\cite{defard2020padim,li2021cutpaste}, or the kNN for the entire normal image embedding~\cite{reiss2021panda,roth2021towards}. We follow the methodology in representation-based method which extract the visual feature from vision transformer or ResNet and establish the distribution through FastFlow model.

\begin{figure*}[t]
\centering
\includegraphics[width=0.99\linewidth]{ 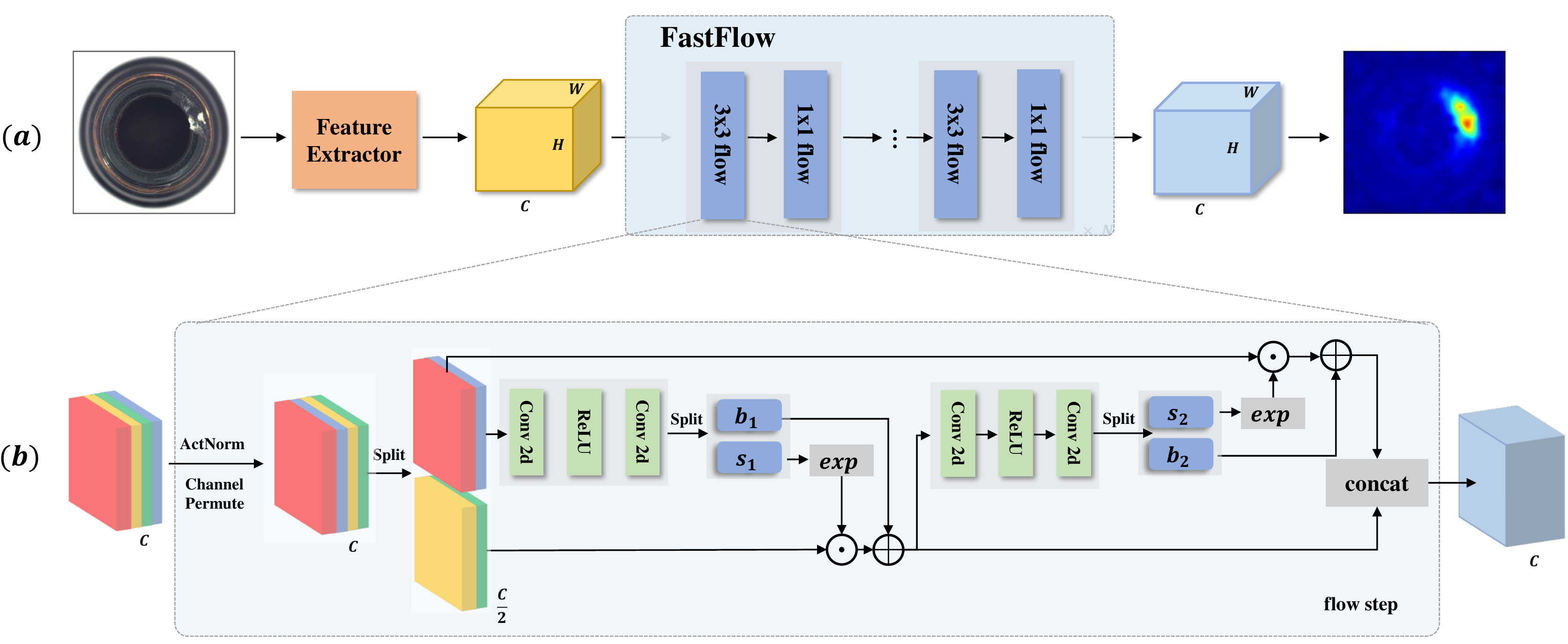} 
\caption{(a) the whole pipeline for unsupervised anomaly detection and localization in our method, which consists of a feature extractor and our FastFlow model. We can use an arbitrary network as the feature extractor such as CNN or vision transformer. FastFlow is alternatly stacked by the ``$3\times3$" and ``$1\times1$" flow.  (b) one flow step for our FastFlow, the ``Conv 2d" can be $3\times3$ or $1\times1$ convolution layer for $3\times3$ or $1\times1$ flow, respectively.} 
\label{fig:pipeline}
\end{figure*}

\subsection{Feature extractors for Anomaly Detection}
With the development of deep learning, recent unsupervised anomaly detection methods use deep neural networks as feature extractors, and produce more promising anomaly results. Most of them~\cite{cohen2020sub, defard2020padim,roth2021towards} use ResNet~\cite{he2016deep} to extract distinguish visual features. Some work has also begun to introduce ViT~\cite{dosovitskiy2020image} into unsupervised anomaly detection fields, such as VT-ADL~\cite{mishra2021vt} uses vision transformer as backbone in a generated-based way. ViT has a global receptive field and can learn the relationship between global and local better. DeiT~\cite{touvron2021training} and CaiT~\cite{touvron2021going} are two typical models for ViT. DeiT introduces a teacher-student strategy specific to transformers, which makes image transformers learn more efficiently and got a new state-of-the-art performance. CaiT proposes a simple yet effective architecture designed in the spirit of encoder/decoder architecture and demonstrates that transformer models offer a competitive alternative to the best convolutional neural networks. In this paper, we use various networks belonging to CNN and ViT to prove the universality of our method.


\subsection{Normalizing Flow}
Normalizing Flows (NF)~\cite{rezende2015variational} are used to learn transformations between data distributions with special property that their transform process is bijective and the flow model can be used in both directions. Real-NVP~\cite{dinh2016density} and Glow~\cite{kingma2018glow} are two typical methods for NF, in which both forward and reverse processes can be processed quickly. NF is generally used to generate data from variables sampled in a specific probability distribution, such as images or audios. Recently, some work~\cite{rudolph2021same,gudovskiy2021cflow} began to use it for unsupervised anomaly detection and localization. DifferNet~\cite{rudolph2021same} achieved good image level anomaly detection performance by using NF to estimate the precise likelihood of test images. Unfortunately, this work failed to obtain the exact anomaly localization results since they flattened the outputs of feature extractor.
CFLOW-AD~\cite{gudovskiy2021cflow} proposes to use hard code position embedding to leverage the distribution learned by NF, which probably underperforms at more complicated datasets.    

\section{Methodology}
In this section, we introduce the pipeline of our method and the architecture of the FastFlow, as shown in Figure~\ref{fig:pipeline}. We first set up the problem definition of unsupervised anomaly detection, and introduce the basic methodology that uses a learnable probability density estimation model in the representation-based method. Then we describe the details of feature extractor and FastFlow models, respectively. 
\subsection{Problem Definition and Basic Methodology}
Unsupervised anomaly detection is also denoted as one-class classification or out-of-distribution detection which requires the model to determines whether the test image is normal or abnormal. Anomaly localization requires a more fine-grained result that gives the anomalies label for each pixel. During the training stage, only normal images were observed, but the normal images and abnormal images simultaneously appear in inference. One of the mainstream methods is representation-based method which extracts discriminative feature vectors from normal images or normal image patches to construct the distribution and calculate anomaly score by the distance between the embedding of a test image and the distribution. The distribution is typically characterized by the center of an n-sphere for the normal image, the Gaussian distribution of normal images, or the normal embedding cluster stored in the memory bank obtained from KNN. After extract the features of the training dataset $D = \{x_1, x_2, \cdots, x_N\}$ where $x_i, i = 1, 2, \cdots, N$ are samples from the distribution $p_{X}(x)$, a representation-based anomaly detection model $\mathcal{P} = \{P_{\theta}:\theta \in \Theta\}$ aims to learn the parameter $\theta$ in the parameter space $\Theta$ to map all $x_{i}$ from the raw distribution $p_{X}(x)$ into the same distribution $p_{Z}(z)$, with anomalous pixels or instances mapped out of the distribution. In our method, we follow this methodology and propose FastFlow $P_{\theta}$ to project the high-dimensional visual features of normal images extracted from typical backbone networks into the standard normal distribution.

\subsection{Feature Extractor}
In the whole pipeline of our method, we first extract the representative feature from the input image through ResNet or vision transformers. As mentioned in the Sec~\ref{section:intro}, one of significant challenges in the anomaly detection task is the global relation grasped to distinguish those abnormal regions from other local parts. Therefore, when using vision transformer (ViT)~\cite{dosovitskiy2020image} as the feature extractor, we only use the feature of one certain layer because ViT has stronger ability to capture the relationship between local patches and the global feature. For ResNet, we directly use the features of the last layer in the first three blocks, and put these features into three corresponding FastFlow model.

\subsection{2D Flow Model}
As shown in Figure~\ref{fig:pipeline}, our 2D flow $f: X \rightarrow Z$ is used to project the image features $x \in p_{X}(x)$ into the hidden variable $z \in p_{Z}(z)$ with a bijective invertible mapping. For this bijection function, the change of the variable formula defines the model distribution on $X$ by:

\begin{equation}
\begin{aligned}
{p}_{X}(x) ={p}_{Z}(z) \left\vert \mathbf{det}(\frac{\partial z}{\partial x})\right\vert
\end{aligned}
\end{equation}

We can estimate the log likelihoods for image features from $p_{Z}(z)$ by:
\begin{equation}
\begin{aligned}
\log {p}_{X}(x) &= \log p_{Z}(z) + \log \left\vert \mathbf{det}(\frac{\partial z}{\partial x}) \right\vert \\
& = \log p_{Z}(f_{\theta}(x)) + \log \left\vert \mathbf{det}(\frac{\partial f_{\theta}(x)}{\partial x}) \right\vert , 
\end{aligned}
\end{equation}
where $z \sim \mathcal{N}(o, I)$ and the $\mathbf{\frac{\partial f_{\theta}(x)}{\partial x}}$ is the Jacobian of a bijective invertible flow model that $z = f_{\theta}(x)$ and $x = f^{-1}_{\theta}(z)$, $\theta$ is parameters of the 2D flow model. 
In inference, the features of anomalous images should be out of distribution and hence have lower likelihoods than normal images and the likelihood can be used as the anomaly score.
Specifically, we sum the two-dimensional probabilities of each channel to get the final probability map and upsample it to the input image resolution using bilinear interpolation.
In actual implementation, our flow model $f_{2d}$ is constructed by stacking multiple invertible transformations blocks $f_{i}$ in a sequence that:
$$
X \xrightarrow{f_{1}} H_{1} \xrightarrow{f_{2}} H_{2} \xrightarrow{f_{3}} \cdots \xrightarrow{f_{K}} Z,
$$
and 
$$
X \xleftarrow{f^{-1}_{1}} H_{1} \xleftarrow{f^{-1}_{2}} H_{2} \xleftarrow{f^{-1}_{3}} \cdots \xleftarrow{f^{-1}_{K}} Z,
$$
where the 2D flow model is $f_{2d}=f_{1}\circ f_{2}\circ f_{3}\circ \cdots \circ f_{K}$ with $K$ transformation blocks. Each transformation block $f_{i}$ consists of multiple steps. Following~\cite{dinh2014nice}, we employ affine coupling layers in each block, and each step is formulated as follow:
\begin{equation}
\begin{aligned}
y_{a}, y_{b} &= \text{split}(y) \\
y'_{a} &= y_{a} \\
y'_{b} &= s(y_{a}) \odot y_{b} + b(y_{a}) \\
y' &= \text{concat}(y'_{a}, y'_{b}),
\end{aligned}
\end{equation}
where $s(y_a)$ and $b(y_a)$ are outputs of two neural networks. The split($\cdot$) and concat($\cdot$) functions perform splitting and concatenation operations along the channel dimension. The two subnets s($\cdot$) and b($\cdot$) are usually implemented as fully connected networks in original normalizing flow model and need to flatten and squeeze the input visual features from 2D to 1D which destroy the spatial position relationship in the feature map. To convert the  original normalizing flow to 2D manner, we adopt two-dimensional convolution layer in the default subnet to reserve spatial information in the flow model and adjust the loss function accordingly. In particular, we adopt a fully convolutional network in which 3$\times$3 convolution and 1$\times$1 convolution appear alternately, which reserves spatial information in the flow model.

\begin{table*}[tbp]
    \centering
    \begin{tabular}{c|ccccc}
        \hline
         Model & FPS & A.d. Time (ms) & A.d. Params (M) & Image-level AUC & Pixel-level AUC \\
         \hline
         CaiT-M48-distilled &  &   &  & \\
         + Patch Core & 2.39 & 107 & 0 & 97.9 & 96.5 \\
         + CFlow &  2.76 &  42 & 10.5 & 97.7& 96.2\\
         + FastFlow & \textbf{3.08} & \textbf{9} & 14.8 & \textbf{99.4} & \textbf{98.5}\\
         \hline
         DeiT-base-distilled &   &  &  & \\
         + Patch Core & 15.45 & 39  & 0 & 96.5 & 97.9\\
         + CFlow & 16.91 &  34 & 10.5 & 95.6 & 97.9\\
         + FastFlow & \textbf{30.14} & \textbf{8} & 14.8 & \textbf{98.7} & \textbf{98.1}\\
         \hline
         ResNet18 &   &   & &  &   \\
         + SPADE & 3.92 & 250 & 0 & - & -\\
         + CFlow &   20.3& 44  & 5.5 & 96.8 & \textbf{98.1} \\
        + FastFlow & \textbf{30.8} & \textbf{27} & 4.9  & \textbf{97.9} & 97.2 \\ 
         \hline
         Wide-ResNet50-2 &  &   &  &  \\
         + SPADE      & 0.67 & 1481& 0 & 96.2 & 96.5 \\
         + Patch Core & 5.88 & 159 & 0 & 99.1 & 98.1 \\
         + CFlow & 14.9 & 56 & 81.6 & 98.3 & \textbf{98.6} \\
         + FastFlow & \textbf{21.8} & \textbf{34} & 41.3 & \textbf{99.3} & 98.1\\
         \hline
    \end{tabular}
    \caption{Complexity comparison in terms of inference speed (FPS), additional inference time (millisecond) and number of additional parameters (M) for various backbones. A.d. Time means the additional inference time and A.d. Parmas is the number of additional parameters compared with backbone network.}
    \label{table:complexity}
\end{table*}

\section{Experiments}

\subsection{Datasets and Metrics}
We evaluate the proposed method on three datasets: MVTec AD~\cite{bergmann2019mvtec}, BTAD~\cite{mishra2021vt} and CIFAR-10~\cite{krizhevsky2009learning}. MVTec AD and BTAD are both industrial anomaly detection datasets with pixel-level annotations, which are used for anomaly detection and localization. CIFAR-10 is built for image classification and we use it to do anomaly detection.  Following the previous works, we choose one of the categories as normal, and the rest as abnormal. The anomalies in these industrial datasets are 
finer than those in CIFAR-10, and the anomalies in CIFAR-10 are more related to the semantic high-level information. For example, the anomalies in MVTec AD are defined as small areas, while the anomalies in CIFAR-10 dataset are defined as different object categories. Under the unsupervised setting, we train our model for each category with its respective normal images and evaluate it in test images that contain both normal and abnormal images. 

The performance of the proposed method and all comparable methods is measured by the area under the receiver operating characteristic curve (AUROC) at image or pixel level. For the detection task, evaluated models are required to output single score (anomaly score) for each input test image. In the localization task, methods need to output anomaly scores for every pixel.

\subsection{Complexity Analysis}
We make a complexity analysis of FastFlow and other methods from aspects of inference speed, additional inference time and additional model parameters, ``additional" refers to not considering the backbone itself. The hardware configuration of the machine used for testing is Intel(R) Xeon(R) CPU E5-2680 V4@2.4GHZ and NVIDIA GeForce GTX 1080Ti. SPADE and Patch Core perform KNN clustering between each test feature of each image patch and the gallery features of normal image patches, and they do not need to introduce parameters other than backbone. CFlow avoids the time-consuming k-nearest-neighbor-search process, but it still needs to perform testing phase in the form of a slice window. Our FastFlow adopts an end-to-end inference phase which has high efficiency of inference. The analysis results are shown in Table~\ref{table:complexity}, we can observe that our method is up to $10\times$ faster than other methods. Compared with CFlow which also uses flow model, our method achieves $1.5\times$ speedup and $2\times$ parameter reduction. When using vision transformers (deit and cait) as the feature extractor, our FastFlow can achieve 99.4 image-level AUC for anomaly detection which is superior to CFlow and Patch Core. From the perspective of additional inference time, our method achieves up to $4\times$ reduction compared to Cflow and $10\times$ reduction compared to Patch Core. Our FastFlow can still have a competitive performance when using ResNet model as feature extractor.

\begin{table*}[tbp]
\begin{center}
\resizebox{\linewidth}{!}{
\begin{tabular}{c|c|c|c|c|c|c|c|c}
\hline
Method & PatchSVDD & SPADE* & DifferNet & PaDiM & Cut Paste & PatchCore & CFlow & FastFlow \\
\hline
carpet           & (92.9,92.6) & (98.6,97.5) & (84.0,-)  & (-,99.1) & (\textbf{100.0},98.3) & (98.7,98.9)  & (\textbf{100.0},99.3)  & (\textbf{100.0},\textbf{99.4})   \\
grid             & (94.6,96.2) & (99.0,93.7) & (97.1,-)  & (-,97.3) & (96.2,97.5) & (98.2,98.7)  & (97.6,\textbf{99.0})  & (\textbf{99.7},98.3)   \\
leather          & (90.9,97.4) & (99.5,97.6) & (99.4,-)  & (-,99.2) & (95.4,99.5) & (\textbf{100.0},99.3)  & (97.7,\textbf{99.7})  & (\textbf{100.0},99.5)   \\
tile             & (97.8,91.4) & (89.8,87.4) & (92.9,-)  & (-,94.1) & (\textbf{100.0},90.5) & (98.7,95.6)  & (98.7,\textbf{98.0})  & (\textbf{100.0},96.3)   \\
wood             & (96.5,90.8) & (95.8,88.5) & (99.8,-)  & (-,94.9) & (99.1,95.5) & (99.2,95.0)  & (99.6,96.7)  & (\textbf{100.0},\textbf{97.0})   \\
bottle           & (98.6,98.1) & (98.1,98.4) & (99.0,-)  & (-,98.3) & (99.9,97.6) & (\textbf{100.0},98.6)  & (\textbf{100.0},\textbf{99.0})  & (\textbf{100.0},97.7)   \\
cable            & (90.3,96.8) & (93.2,97.2) & (86.9,-)  & (-,96.7) & (\textbf{100.0},90.0) & (99.5,\textbf{98.4})  & (\textbf{100.0},97.6)  & (\textbf{100.0},\textbf{98.4})   \\
capsule          & (76.7 ,95.8) & (98.6,99.0) & (88.8,-)  & (-   ,98.5) & (98.6 ,97.4) & (98.1,98.8)  & (99.3 ,99.0)  & (\textbf{100.0},\textbf{99.1})   \\
hazelnut         & (92.0 ,97.5) & (98.9,99.1) & (99.1,-)  & (-   ,98.2) & (93.3 ,97.3) & (\textbf{100.0},98.7)  & (96.8 ,98.9)  & (\textbf{100.0},\textbf{99.1})   \\
meta nut         & (94.0,98.0) & (96.9,98.1) & (95.1,-)  & (-,97.2) & (86.6,93.1) & (\textbf{100.0},98.4)  & (91.9,\textbf{98.6})  & (\textbf{100.0},98.5)   \\
pill             & (86.1,95.1) & (96.5,96.5) & (95.9,-)  & (-,95.7) & (\textbf{99.8},95.7) & (96.6,97.1)  & (99.9,99.0)  & (99.4,\textbf{99.2})   \\
screw            & (81.3,95.7) & (99.5,98.9) & (99.3,-)  & (-,98.5) & (90.7,96.7) & (98.1,\textbf{99.4})  & (\textbf{99.7},98.9)  & (97.8,\textbf{99.4})   \\
toothbrush       & (\textbf{100.0},98.1) & (98.9,97.9) & (96.1,-)  & (-,98.8) & (97.5 ,98.1) & (\textbf{100.0},98.7)  & (95.2,\textbf{99.0})  & (94.4 ,98.9)   \\
transistor       & (91.5,97.0) & (81.0,94.1) & (96.3,-)  & (-,97.5) & (99.8,93.0) & (\textbf{100.0},96.3)  & (99.1 ,\textbf{98.0})  & (99.8,97.3)   \\
zipper           & (97.9,95.1) & (98.8,96.5) & (98.6,-)  & (-,98.5) & (\textbf{99.9},\textbf{99.3}) & (98.8,98.8)  & (98.5 ,99.1) & (99.5,98.7)   \\
\hline
 AUCROC  & (92.1,95.7) & (96.2,96.5) & (94.9,-)  & (97.9,97.5) & (97.1,96.0) & (99.1,98.1)  & (98.3,\textbf{98.6})  & (\textbf{99.4},98.5) \\
 \hline
\end{tabular}
}
\caption{Anomaly detection and localization performance on MVTec AD dataset with the format (image-level AUC, pixel-level AUC). We report the detailed results for all categories.}
\label{tab:image-level}
\end{center}
\end{table*}

\subsection{Quantitative Results}
\subsubsection{MVTec AD} There are 15 industrial products in MVTec AD dataset~\cite{bergmann2019mvtec}, with a total of 5,354 images, among which 10 are objects and the remaining 5 are textures. The training set is only composed of normal images, while the test set is a mixture of normal images and abnormal images. We compare our proposed method with the state-of-the-art anomaly detection works, including SPADE*~\cite{reiss2021panda}, PatchSVDD~\cite{yi2020patch}, DifferNet~\cite{rudolph2021same}, Mah.AD~\cite{rippel2021modeling}, PaDiM~\cite{defard2020padim}, Cut Paste~\cite{li2021cutpaste}, Patch Core~\cite{roth2021towards}, CFlow~\cite{gudovskiy2021cflow} under the metrics of image-level AUC and pixel-level AUC. The detailed comparison results of all categories are shown in Table~\ref{tab:image-level}. We can observe that FastFlow achieves \textbf{99.4} AUC on image-level and \textbf{98.5} AUC on pixel-level, suppresses all other methods in anomaly detection task.


\begin{table}[tbp]
    \centering
    \resizebox{\linewidth}{!}{
    \begin{tabular}{c|c|c|c|c}
         Categories &  AE MSE & AE MSE+SSIM & VT-ADL & FastFlow \\
         \hline
         0 & 0.49 & 0.53 & 0.99 & 0.95 \\
         \hline
         1 & 0.92 & 0.96 & 0.94 & 0.96 \\
         \hline
         2 & 0.95 & 0.89 & 0.77 & 0.99 \\
         \hline
         Mean & 0.78 & 0.79 & 0.90 & \textbf{0.97}\\
         \hline
    \end{tabular}}
    \caption{Anomaly localization results on BTAD datasts. We compare our method with convolutional auto encoders trained with MSE-loss and MSE+SSIM loss, and VT-ADL.}
    \label{table:btad}
\end{table}

\subsubsection{BTAD} BeanTech Anomaly Detection dataset~\cite{mishra2021vt} has 3 categories industrial products with 2540 images. The training set consists only of normal images, while the test set is a mixture of normal images and abnormal images. Under the measure of pixel-level AUC, we compare the results of our FastFlow with the results of three methods reported in VT-ADL~\cite{mishra2021vt}: auto encoder with mean square error, automatic encoder with SSIM loss and VT-ADL. The comparison results are shown in Table~\ref{table:btad}. We can observe that our FastFlow achieves 97.0 pixel-wise AUC and suppresses other methods as high as 7\% AUC.

\begin{table}[tbp]
    \centering
    \resizebox{\linewidth}{!}{
    \begin{tabular}{c|c|c|c|c|c}
        \hline
        Method & OC-SVM & KDE & $\textit{l}_{2}$-AE & VAE & Pixel CNN \\
        \hline
        AUC & 58.6 & 61.0 & 53.6 & 58.3 & 55.1  \\
        \hline
        Method & LSA & AnoGAN & DSVDD & OCGAN & FastFlow \\
        \hline
         AUC &  64.1 & 61.8 & 64.8 & 65.6 & \textbf{66.7} \\
        \hline 
    \end{tabular}}
    \caption{Anomaly detection results on CIFAR-10 datast.}
    \label{tab:cifar10}
\end{table}

\subsubsection{CIFAR-10 dataset} CIFAR-10 has 10 categories with 60000 natural images. Under the setting of anomaly detection, one category is regarded as anomaly and other categories are used as normal data. And we need to train the corresponding model for each class respectively. The AUC scores of our method and other methods are reported in Table~\ref{tab:cifar10}. Methods for comparison includes OC-SVM~\cite{scholkopf1999support}, KDE~\cite{bishop2006pattern}, $\textit{l}_{2}$-AE~\cite{hadsell2006dimensionality}, VAE~\cite{an2015variational}, Pixel CNN~\cite{oord2016conditional}, LSA~\cite{abati2019latent}, AnoGAN~\cite{schlegl2017unsupervised}, DSVDD~\cite{ruff2018deep} and OCGAN~\cite{perera2019ocgan}. Our method outperforms these comparison methods. The results in three different datasets show that our method can adapt to different anomaly detection settings.



\subsection{Ablation Study}
To investigate the effectiveness of the proposed FastFlow structure, we design ablation experiments about the convolution kernel selection in subnet. We compare alternately using  $3\times3$ and $1\times1$ convolution kernel and only using $3\times3$ kernel under the AUC and inference speed for the subnet with various backbone networks. The results are shown in Table~\ref{table:ablation}. For the backbone network with large model capacities such as CaiT and Wide-ResNet50-2, alternate using $3\times3$ and $1\times1$ convolution layer can obtain higher performance while reducing the amount of parameters. For the backbone network with small model capacities such as DeiT and ResNet18, only using $3\times3$ convolution layer has higher performance. To achieve the balance of accuracy and inference speed, we use alternate convolution kernels of $3\times3$, $1\times1$ with DeiT, CaiT and Wide-ResNet50-2, and only use $3\times3$ convolution layer with ResNet18.

\begin{figure*}[t]
\centering
\includegraphics[width=0.98\linewidth]{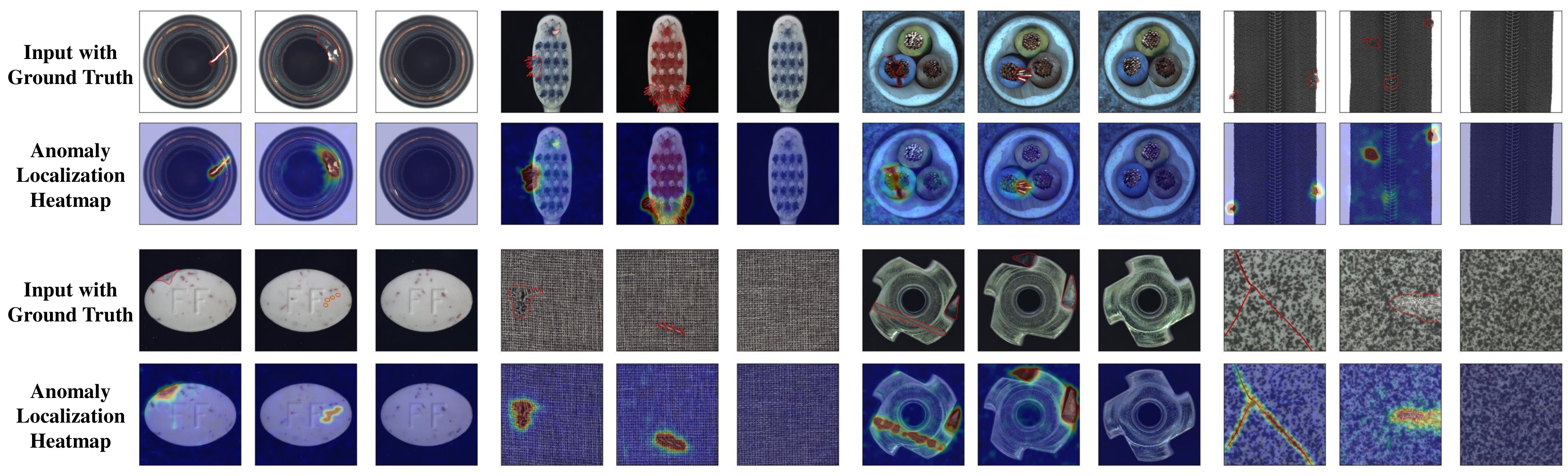} 
\caption{Anomaly localization results of MVTec AD datasets. From top to bottom, input images with ground-truth localization area labeled in red and anomaly localization heatmaps.}
\label{fig:results_vis}
\end{figure*}

\begin{figure}[t]
\centering
\includegraphics[width=0.975\columnwidth]{ 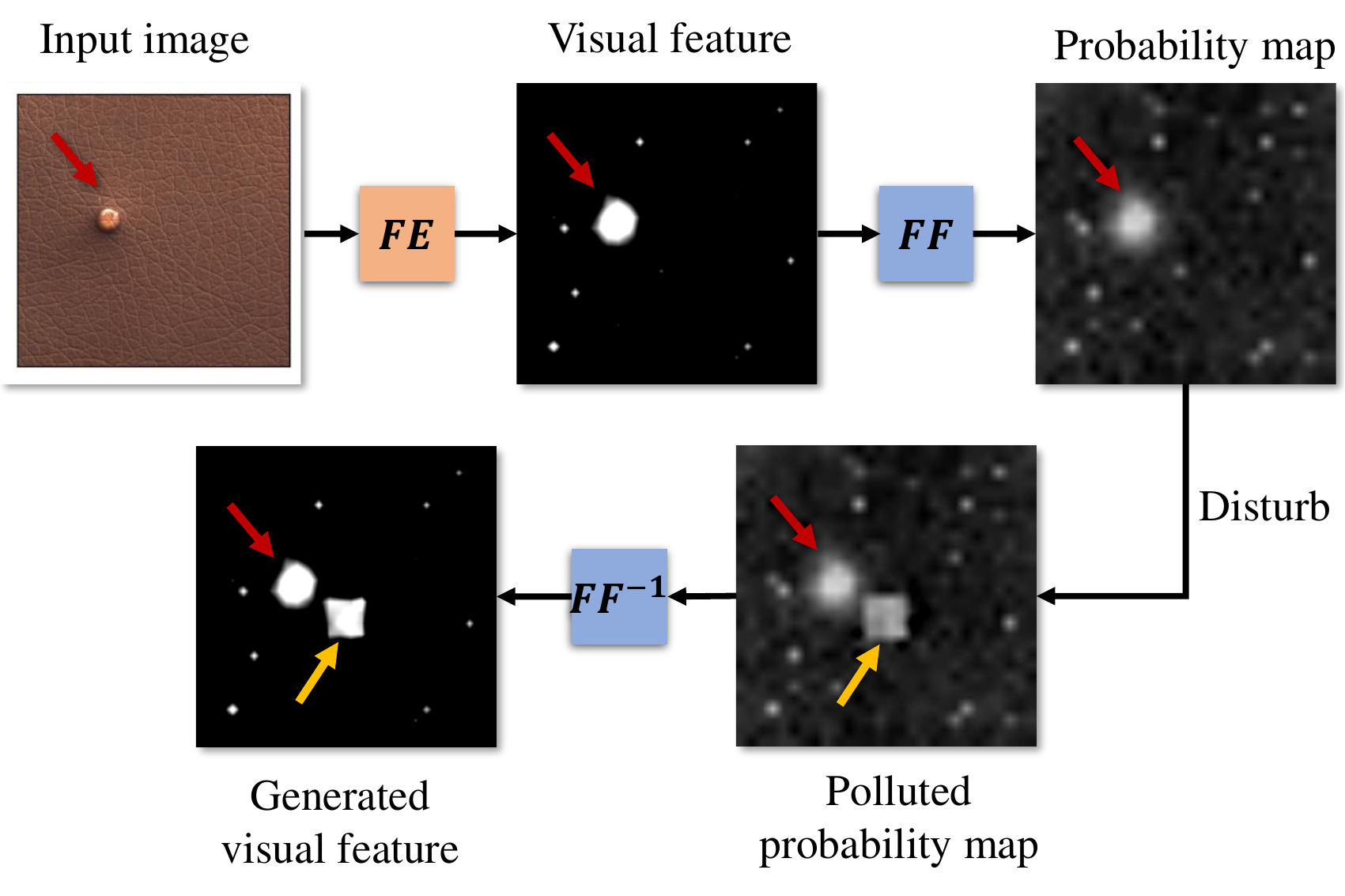} 
\caption{The bidirectional invertible process for FastFlow. ``FE" is the feature extractor, ``FF" is our FastFlow model, ``$\text{FF}^{-1}$" is the reverse for FastFlow. The red and yellow arrows point to the original anomaly and the new anomaly introduced after the noise disturbance respectively.}
\label{fig:vis_flow}
\end{figure}

\subsection{Feature Visualization and Generation.}
Our FastFlow model is a bidirectional invertible probability distribution transformer. In the forward process, it takes the feature map from the backbone network as input and transforms its original distribution into a standard normal distribution in two-dimensional space. In the reverse process, the inverse of FastFlow can generate the visual feature from a specific probability sampling variable. To better understand this ability in view of our FastFlow, we visualize the forward (from visual features to probability map) and reverse (from probability map to visual features) processes. As shown in Figure~\ref{fig:vis_flow}, we extract the features of an input image belonging to the leather class and the abnormal area is indicated by the red arrow. We forward it through the FastFlow model to obtain the probability map. Our FastFlow successfully transformed the original distribution into the standard normal distribution. Then, we add noise interference to a certain spatial area which is indicated by the yellow arrow in this probability map, and generate a leather feature tensor from the pollution probability map by using the inverse Fastflow model. In which we visualized the feature map of one channel in this feature tensor, and we can observe that new anomaly appeared in the corresponding pollution position. 

\begin{table}[tbp]
    \centering
    \resizebox{\linewidth}{!}{
    \begin{tabular}{c|ccc}
    Method & A.d. Params (M) & Image-level AUC & Pixel-level AUC \\
    \hline
    DeiT & \\
    3-1 & 14.8 & 98.7 & 98.1\\
    3-3 & 26.6 & 98.7 & \textbf{98.3} \\
    \hline
    CaiT & \\
    3-1 & 14.8 & \textbf{99.4} & 98.5 \\
    3-3 & 26.6 & 98.9 & 98.5 \\
    \hline
    ResNet18 & \\
    3-1 & 2.7 & 97.3 & 96.8 \\
    3-3 & 4.9 & \textbf{97.9} & \textbf{97.2} \\
    \hline
     Wide-ResNet50-2 & \\
    3-1 & 41.3 & \textbf{99.3} & \textbf{98.1} \\
    3-3 & 74.4 & 98.2 & 97.6 \\
    \hline
    \end{tabular}}
    \caption{Results of ablation experiments with various backbone networks. 3-1 means alternately using $3\times 3$ and $1\times 1$ convolution layers and 3-3 is only using $3\times 3$ convolution layer in the subnet for FastFlow. A.d. Params is the number of additional model parameters compared with backbone network.}
    \label{table:ablation}
\end{table}

\subsection{Qualitative Results}
We visualize some results of anomaly detection and localization in Figure~\ref{fig:results_vis} with the MVTec AD dataset. The top row shows test images with ground truth masks with and without anomalies, and the anomaly localization score heatmap is shown in the bottom row. There are both normal and abnormal images and our FastFlow gives accurate anomaly localization results.

\begin{table}[t]
    \centering
    \resizebox{\linewidth}{!}{
    \begin{tabular}{c|c|c|c}
        \hline
         Backbone & Input Size & Block Index & Feature Size\\
         \hline
         CaiT-M48-distilled & 448 & 40 & 28  \\
         \hline
         DeiT-base-distilled & 384 & 7 & 24  \\
         \hline
         Res18 & 256 & [1,2,3] & [64, 32, 16]   \\
         \hline
         WR50 & 256 & [1,2,3] & [64, 32, 16] \\
         \hline
    \end{tabular}}
    \caption{We use four different feature extractors in all experiments. The input picture size and feature size are set according to the backbone network and the block index indicates the block from which the feature is obtained..}
    \label{table:implementation}
\end{table}

\subsection{Implementation Details}
We provide the details of the structure of feature extractor, the selection of feature layer and the size of input image in Table~\ref{table:implementation}. For vision transformer, our method only uses feature maps of a specific layer, and does not need to design complicated multi-scale features manually. For ResNet18 and Wide-ResNet50-2, we directly use the features of the last layer in the first three blocks, put these features into the 2D flow model to obtain their respective anomaly detection and localization results, and finally take the average value as the final result. All these backbone are initialized with the ImageNet pre-trained weights and their parameters are frozen in the following training process. For FastFlow, we use 20-step flows in CaiT and DeiT and 8-step flows for ResNet18 and Wide-ResNet50-2. We train our model using Adam optimizer with the learning rate of 1e-3 and weight decay of 1e-5. We use a 500 epoch training schedule, and the batch size is 32.

\section{Conclusion}
In this paper, we propose a novel approach named FastFlow for unsupervised anomaly detection and localization. Our key observation is that anomaly detection and localization requires comprehensive consideration of global and local information with a learnable distribution modeling method, and efficient inference process, which are ignored in the existing approaches. To this end, we present a 2D flow model denoted as FastFlow which has a lightweight structure and is used to project the feature distribution of normal images to the standard normal distribution in training, and use the probabilities as the anomaly score in testing. FastFlow can be used in typical feature extraction networks such as ResNet and ViT in the form of plug-ins.  Extensive experimental results on MVTec AD dataset show FastFlow superiority over the state-of-the art methods in terms of accuracy and reasoning efficiency.


\clearpage

\twocolumn[
\begin{@twocolumnfalse}
	\section*{\centering{Supplementary Material for \\ \emph{FastFlow: Unsupervised Anomaly Detection and Localization via 2D Normalizing Flows\\[25pt]}}}
\end{@twocolumnfalse}
]

\section{More Ablation Studies}

\subsection{Channels of Hidden Layers in Flow Model}
In the original flow model which has been used in DifferNet~\cite{rudolph2021same} and CFLOW~\cite{gudovskiy2021cflow},
the number of channels of hidden layers in all subnet is set to $2\times$ as much the input and output layer's channel. This kind of design improves the results by increasing the complexity of the model, but it reduces the efficiency of inference. In our FastFlow, we found that using $0.16\times$ number of channels in CaiT and $1\times$ number of channels in Wide-ResNet50-2 can achieve a balance between performance and model parameters. In addition, when we use $0.25\times$ number of channels of Wide-ResNet50-2, we can further reduce the model parameters while still maintaining high accuracy. The results are shown in Table~\ref{tab:alb}.

\begin{table}[]
    \centering
    \resizebox{\linewidth}{!}{
    \begin{tabular}{c|ccc}
    \hline
         Channel Ratio & Parameters (M) & Image-level AUC & Pixel-level AUC \\
        \hline
        \hline
        CaiT  \\
        \hline
        $0.16\times$ & 14.8 & 99.4 & 98.5 \\
        $0.33\times$ & 29.6 & 98.9 & 98.4 \\
        \hline
        Wide-ResNet50-2  \\
        \hline
        $0.25\times$ & 10.9 & 98.9 & 98.0 \\
        $0.5\times$ & 20.7 & 99.1 & 98.1 \\
        $1.0\times$ & 41.3 & 99.3 & 98.1 \\
        $2.0\times$ & 82.6 & 99.4 & 98.1 \\
        \hline
    \hline
    \end{tabular}}
    \caption{Ablation study results about the hidden layer channels for CNN and vision transformer in MVTec AD dataset. Channel Ratio means the ratio of the number of channels in the hidden layer to the number of channels in the input and output layers for subnet in our FastFlow.}
    \label{tab:alb}
\end{table}

\subsection{Training Data Augmentation}
In order to learn a more robust FastFlow model, we apply various data augmentation methods to the MVTec AD dataset during the training phase.  We use random horizontal flip, vertical flip and rotation, with probabilities of 0.5, 0.3 and 0.7, respectively. It should be noted that some categories are not suitable for violent data augmentation. For example, the transistor can not be flipped upside down and rotated. 
The results are shown in Table~\ref{tab:data_aug}.

\begin{table}[t]
    \centering
    \begin{tabular}{c|cc}
    \hline
         Data Augmentation & Image-level AUC & Pixel-level AUC \\
        \hline
        CaiT \\
        w/o & 99.3 & 98.4 \\
        w & 99.4 & 98.5 \\
        \hline
        Wide-ResNet50-2 \\
        w/o & 98.9 & 98.2\\
        w & 99.3 & 98.1 \\ 
    \hline
    \end{tabular}
    \caption{The effect of data augmentation on the anomaly detection and localization performance.}
    \label{tab:data_aug}
\end{table}


\section{Bad Cases and Ambiguity Label}
\begin{figure*}[t]
\centering
\includegraphics[width=0.9\linewidth]{ 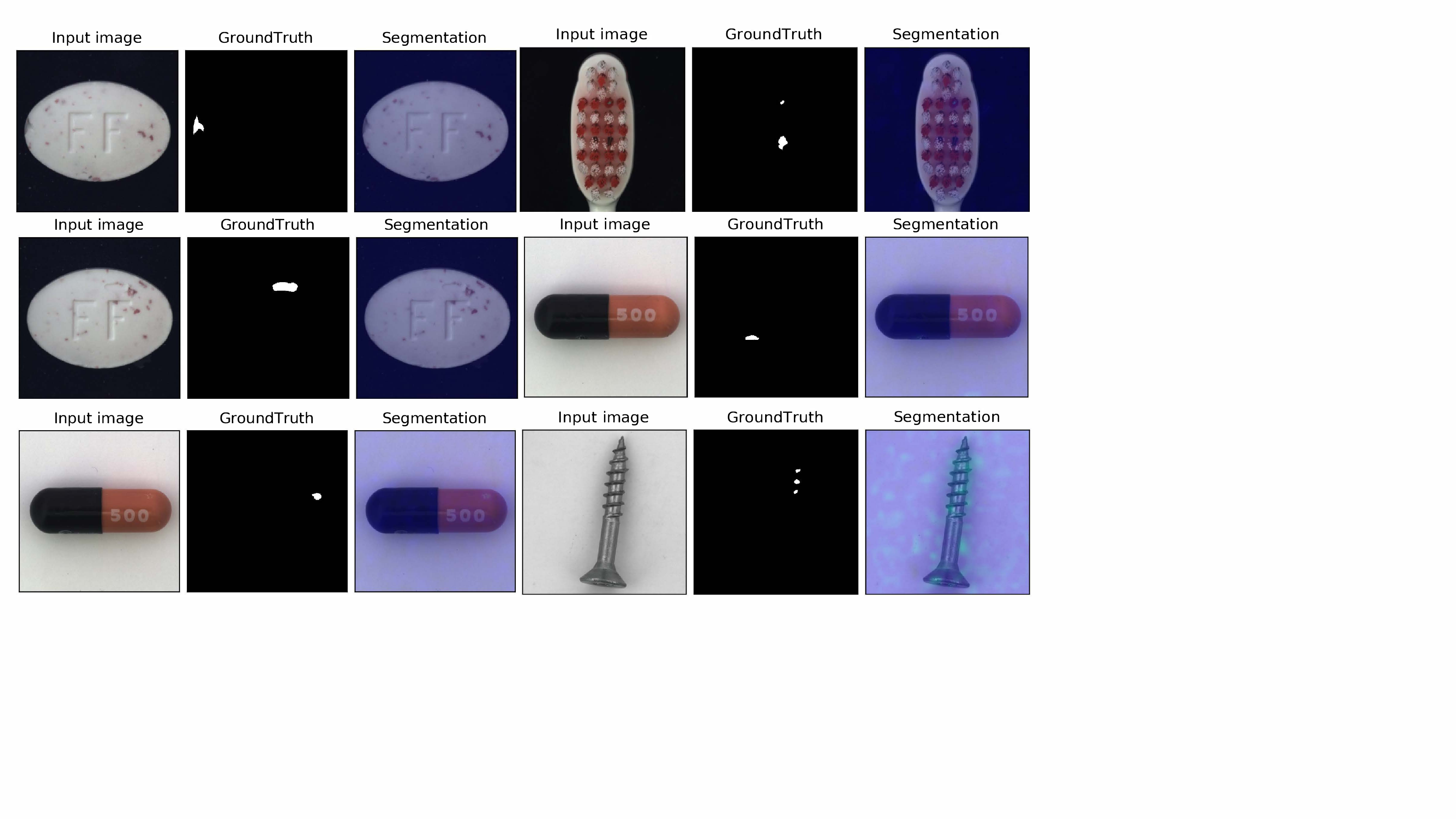} 
\caption{Bad cases of the missing detection type. All missed detection results of our method in shown in this figure.}
\label{fig:fig1}
\end{figure*}

\begin{figure*}[t]
\centering
\includegraphics[width=0.9\linewidth]{ 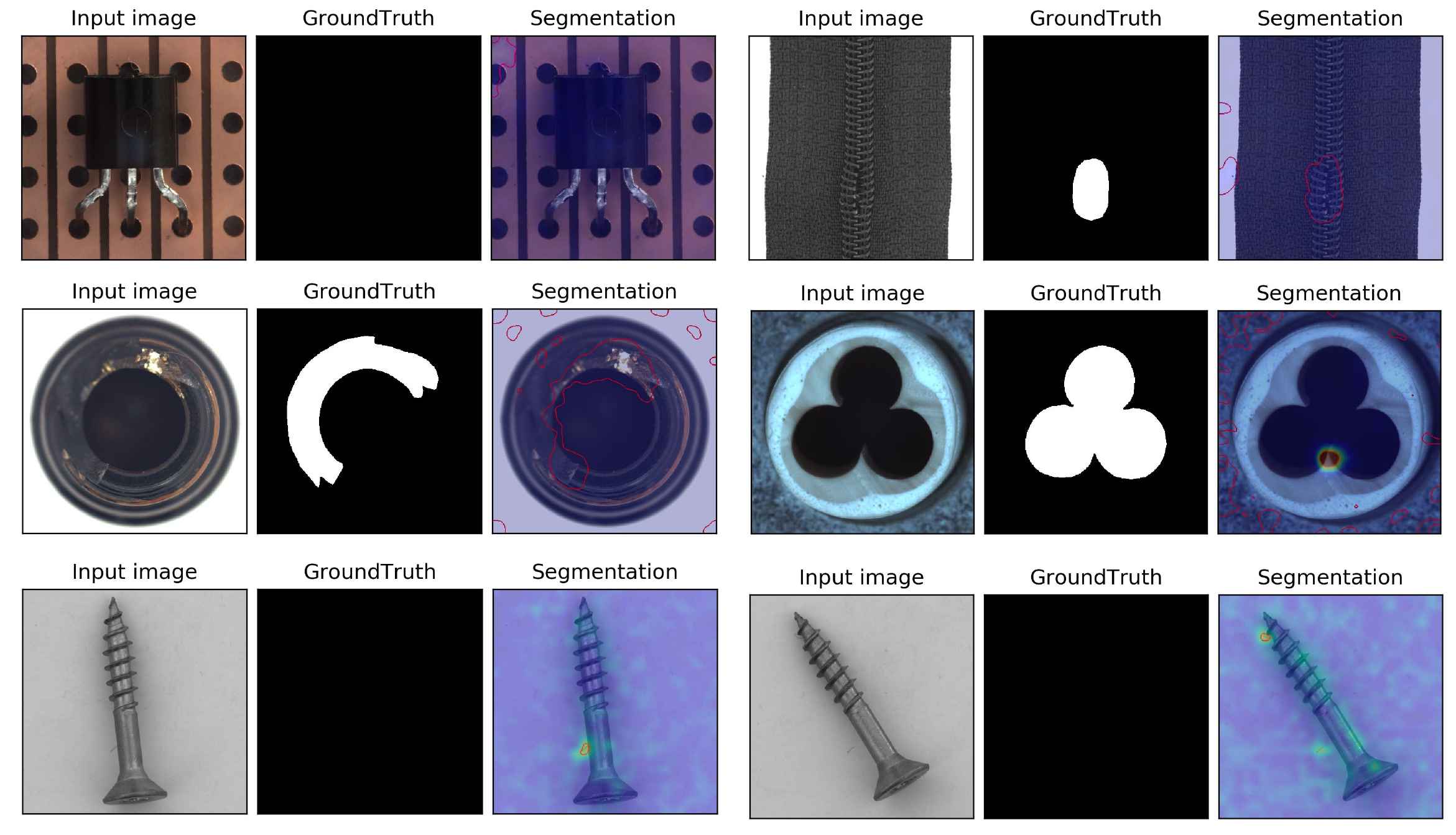} 
\caption{Bad cases of false detection type. We give the typical results of our method in this figure.}
\label{fig:fig2}
\end{figure*}

\begin{figure*}[t]
\centering
\includegraphics[width=0.9\linewidth]{ 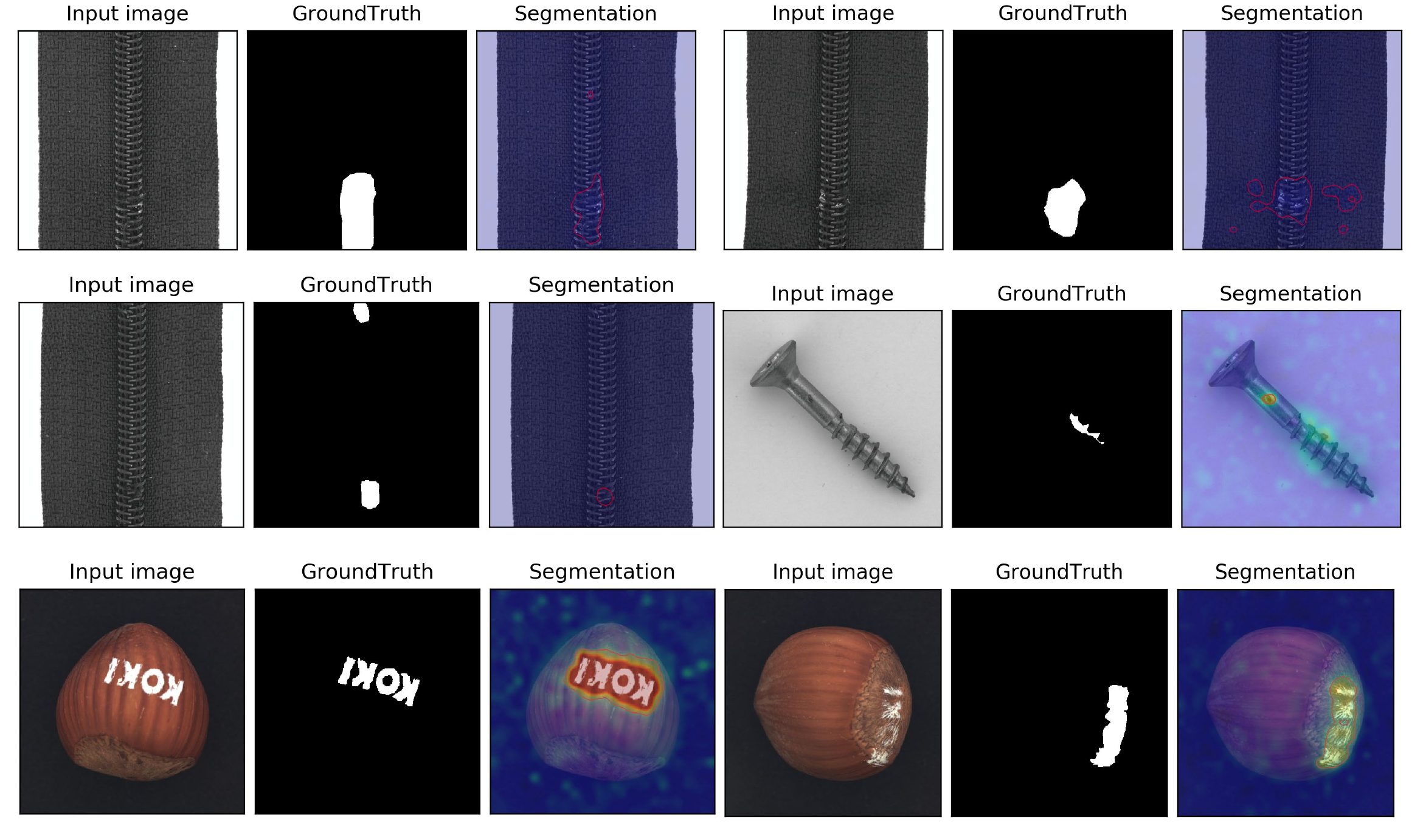} 
\caption{Bad cases caused by label ambiguity. In the first two rows, there are abnormal areas localized by our method while not labeled. In the last row of hazelnut, we show the label ambiguity of the ``print" subclass, in which one hazelnut print is labeled finely, while the other is labeled with a rough area.}
\label{fig:fig3}
\end{figure*}

We visualize bad cases for our method on MVTec AD dataset in Figure~\ref{fig:fig1} to Figure~\ref{fig:fig3} which are summarized into three categories. We show the missing detection cases in Figure~\ref{fig:fig1}, false detection cases in Figure~\ref{fig:fig2} and label ambiguity cases in Figure~\ref{fig:fig3}. In Figure~\ref{fig:fig1}, our method missed a few small and unobvious anomalies. In Figure~\ref{fig:fig2}, our method had false detection results in some background areas, such as areas with hair and dirt in the background. In Figure~\ref{fig:fig3}, our method found some areas belong to abnormal but not be labeled, such as the ``scratch neck" for screw and the ``fabric interior" for zipper.

\section{Non-aligned Disturbed MVTec AD Dataset}
\begin{figure*}[t]
\centering
\includegraphics[width=0.83\linewidth]{ 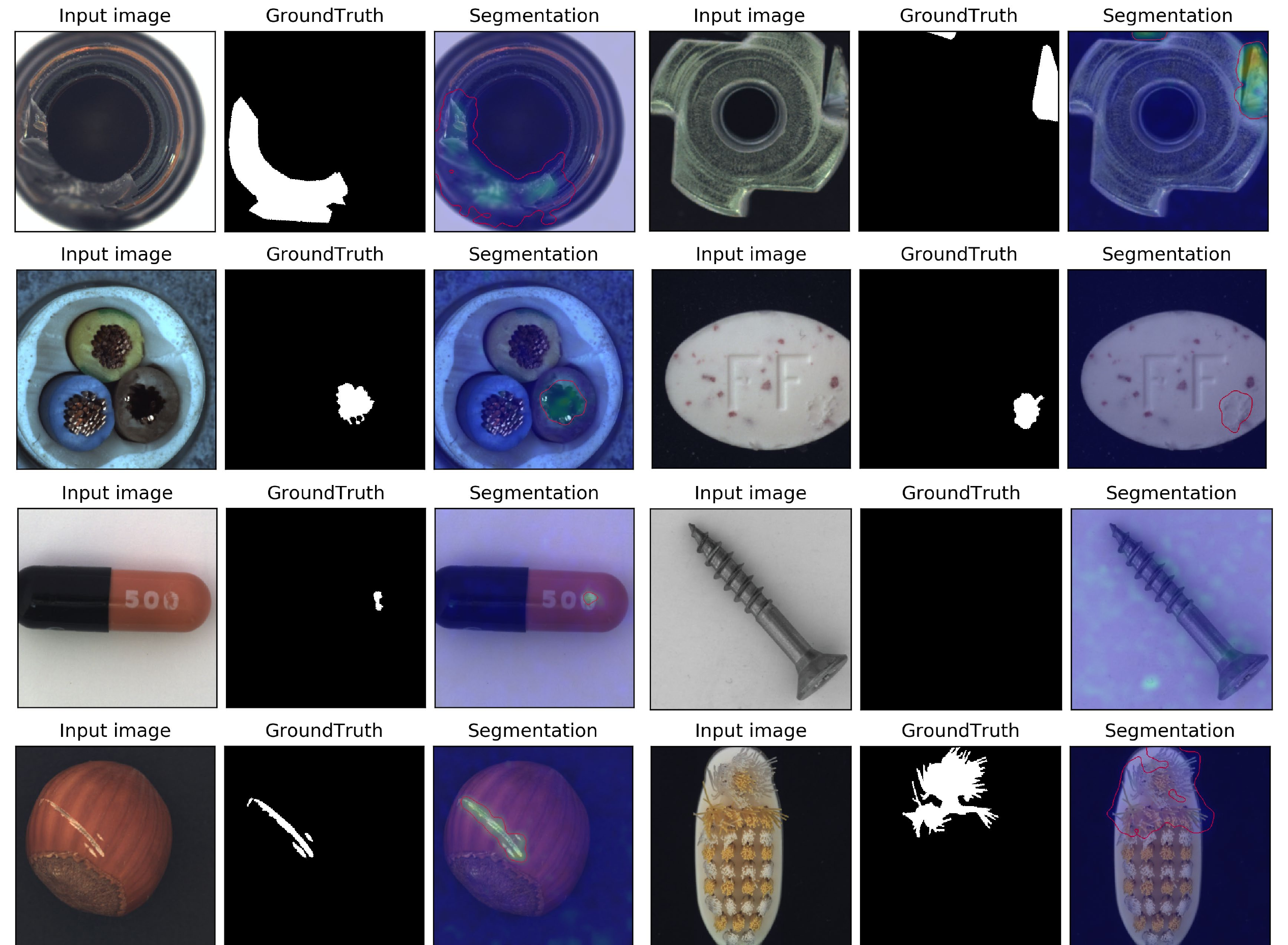} 
\caption{Anomaly localization results of the non-aligned disturbed MVTec AD datasets.}
\label{fig:fig4}
\end{figure*}

Considering that the MVTec AD dataset has the characteristic of sample alignment which is infrequent in practical application, we perform a series of spatial perturbations on the test data to obtain an unaligned MVTec AD dataset. In detail, we apply random zoom in/out with 0.85 ratio, random rotation with $\pm15$ angle, random translation with 0.15 ratio to expand the original test dataset by $4\times$ to the new test dataset. 
We evaluate our FastFlow (with CaiT) in this new test dataset and we obtain 99.2 image-level AUC and 98.1 pixle-level AUC. There is almost no performance loss compared with the results in original aligned MVTec AD test dataset, which proves the robustness of our method. We also give some visualization results in Figure~\ref{fig:fig4}. We can observe that FastFlow can still have high performance on anomaly detection and location result in this non-aligned disturbed MVTec AD dataset.

\bibliography{aaai22}




\end{document}